\documentclass[runningheads]{llncs}
\usepackage{graphicx}

\usepackage{tikz}
\usepackage{comment}
\usepackage{amsmath,amssymb} 
\usepackage{color}
\usepackage{epsfig}
\usepackage{graphicx}
\usepackage{amsmath}
\usepackage{amssymb}
\usepackage{paralist}
\usepackage{xcolor}
\usepackage{booktabs}  
\usepackage{multirow} 
\usepackage{multicol} 
\usepackage{arydshln}
\usepackage{subfigure}

\usepackage{enumitem}
\usepackage{algorithm}
\usepackage{algorithmic}
\usepackage{verbatim}

\newsavebox\CBox
\def\textBF#1{\sbox\CBox{#1}\resizebox{\wd\CBox}{\ht\CBox}{\textbf{#1}}}

\usepackage[accsupp]{axessibility}  

\begin{document}
\pagestyle{headings}
\mainmatter
\def\ECCVSubNumber{5211}  

\setlength{\abovedisplayskip}{4pt} 
\setlength{\belowdisplayskip}{4pt}
\setlength{\abovecaptionskip}{4pt}

\title{Continual Variational Autoencoder Learning via Online Cooperative Memorization} 

\titlerunning{Online Cooperative Memorization for VAE}
%
\author{Fei Ye and Adrian G. Bors}
\authorrunning{Fei Ye and Adrian G. Bors}
%
\institute{Department of Computer Science, University of York, York YO10 5GH, UK\\
\email{\{fy689,adrian.bors\}@york.ac.uk}}
\maketitle

\begin{abstract}
Due to their inference, data representation and reconstruction properties, Variational Autoencoders (VAE) have been successfully used in continual learning classification tasks. However, their ability to generate images with specifications corresponding to the classes and databases learned during Continual Learning (CL) is not well understood and catastrophic forgetting remains a significant challenge. In this paper, we firstly analyze the forgetting behaviour of VAEs by developing a new theoretical framework that formulates CL as a dynamic optimal transport problem. This framework proves approximate bounds to the data likelihood without requiring the task information and explains how the prior knowledge is lost during the training process. We then propose a novel memory buffering approach, namely the Online Cooperative Memorization (OCM) framework, which consists of a Short-Term Memory (STM) that continually stores recent samples to provide future information for the model, and a Long-Term Memory (LTM) aiming to preserve a wide diversity of samples. The proposed OCM transfers certain samples from STM to LTM according to the information diversity selection criterion without requiring any supervised signals. The OCM framework is then combined with a dynamic VAE expansion mixture network for further enhancing its performance.
\end{abstract}

\section{Introduction}
\label{sec:intro}
One desired capability for an artificial intelligence system is to continually learn novel concepts without forgetting the knowledge learnt in the past. However, existing artificial systems are far away from such capabilities, characteristic of living organisms. A deep learning model which can recover the training data from a low-dimensional latent code space is the  Variational Autoencoder (VAE) \cite{VAE}. VAEs have been widely used in image synthesis \cite{MixtureOfVAEs,InfoVAEGAN_conference}, semi-supervised learning \cite{InfVAE,JontLatentVAEs} and for image-to-image translation \cite{ItoI_network}. However, similar to other deep learning systems, VAEs suffer from degenerated performance when it is trained successively with new tasks, which is a result of catastrophic forgetting \cite{LifeLong_review}.

Existing works to relieve VAE's forgetting can be summarized as two categories. The first would usually train a generator \cite{Lifelong_VAE,GenerativeLifelong,Generative_replay}, or store a few past learnt samples \cite{GradientEpisodic} in a memory buffer which replays old samples together with learning new tasks to optimize the model. The methods from the second category would focus on dynamically adding new VAE components into a mixture model to adapt to the data distribution shift \cite{NeuralDirichelt,LifelongUnsupervisedVAE} in which prior knowledge is preserved in the frozen network parameters and structures. These approaches have been extended for the case when the model is trained on non-stationary data streams without knowing the task information, a mechanism called Task Free Continual Learning (TFCL) \cite{OnlineContinualLearning,taskFree_CL}. However, the theoretical analysis for VAE's forgetting behaviour under TFCL has not been studied before.

In recent years, some studies have provided the theoretical analysis for continual learning from different perspectives including the NP-hard problem \cite{OptimalCL}, risk bound \cite{LifelongVAEGAN,LifelongInfinite}, Teacher-Student framework \cite{CLTeacherStudent,LifelongTeacherStudent} and game theory \cite{CL_TradeOff}. However, all these approaches require strong assumptions such as clearly defining the task identities, which is not applicable when the task information is missing. In this paper, we bridge this gap by developing a new theoretical framework which formulates TFCL as a dynamic optimal transport (OT) problem, and derives the approximate bounds on the data likelihood. The motivation behind OT is twofold~: \begin{inparaenum}[1)]
\item OT models evaluate distances between pairs of probability density functions \cite{WVAE2} and can be used for deriving the approximate bound to the data likelihood (See Section~\ref{sec:theory}); \item OT can be estimated by employing sampling \cite{GAN}, which is suitable for analysis and verification\end{inparaenum}. The proposed theoretical analysis also highlights that the sample diversity in the memory used for training is crucial for overcoming forgetting and would not require the category information.

Another contribution of this study, inspired by the above mentioned theoretical analysis, is to develop a new memorization approach aiming to store diverse samples for training a VAE through the TFCL. Other approaches have proposed diversifying the information for memorization by evaluating the similarity on the gradient information \cite{GradientLifelong} or by assigning balanced samples to memory buffers according to their categories' information \cite{RainbowMemory,ContinualPrototype}. However, most of these prior approaches require to access supervised signals, which are not available in unsupervised learning. Additionally, these approaches do not have theoretical guarantees and also ignore the data stream future information in the sample selection. Knowing both the past and future information was shown to improve time series prediction \cite{LSTM_Time} and would be helpful for the sample selection. 

In this paper, we address the aforementioned problems by~:
\begin{inparaenum}[1)]
\item Proposing a new learning paradigm called Online Cooperative Memorization (OCM) which consists of three components: a Long-Term Memory (LTM), a Short-Term Memory (STM) and a model (Learner). OCM implements a memorization mechanism which transfers the temporary information from the STM to LTM, according to a certain criterion. 
\item A kernel-based information importance criterion for evaluating the similarity among the data stored in the STM for selecting diverse characteristic samples for LTM, without requiring a class label.
\end{inparaenum}
The kernel evaluation of the similarity of a pair of data samples \cite{KernelMethods}, defined as an inner product of the latent representations of each pair of the data stored in the memory, is shown to be efficient. This procedure ensures achieving an appropriate diversification among the samples stored in the LTM. We summarize our contributions as follows~:
\begin{inparaenum}[1)]
\item Our work is the first to provide theory insights for the forgetting behaviour of VAE under TFCL.
\item We propose the Online Cooperative Memorization (OCM) that can be used in any VAE variant with minimal modification and can also be extended to a dynamic expansion mixture approach to further enhance performance.
\item We propose a new sample selection approach for dynamically transferring selected samples from the STM to LTM without requiring any supervised signal. To our best knowledge, this is the first work to explore the kernel-based distance for the sample selection under TFCL.
\item The proposed sample selection approach can be used in both supervised and unsupervised learning without modifying the selection strategy.
\end{inparaenum}

\section{Related work}

\noindent \textBF{Continual learning.} 
One of the most popular approaches is to use a regularization loss within the optimization procedure \cite{BooVAE,Distilling_nets,LessForgetting,EWC,Lwf,VCL,LifeLong_combination,LayerwiseCL,NullSpaceCL}, where the network parameters which are important to the past learnt data are re-weighted when learning a new task, in order to attempt to preserve past knowledge. Other approaches would employ a small buffer to store a few past data \cite{GradientLifelong,TinyLifelong,FunctionalRegularisation}  or would train a generator as a generative replay network that provides pseudo data samples for the future task learning \cite{Lifelong_VAE,GenerativeLifelong,LifelongUnsupervisedVAE,Generative_replay,LifelongTeacherStudent,LifelongVAEGAN,Lifelonginterpretable,LifelongTwin,LifelongGAN}. However, these approaches can not guarantee the optimal performance on the past task since stored or generated samples can not represent the true underlying data distributions \cite{LifelongInfinite}. This issue can be solved by storing the information of past samples into the network's parameters which are then frozen when learning novel tasks \cite{NeuralDirichelt,LifelongInfinite,LifelongMixuteOfVAEs,Learning_Evolved,LifelongGraph}. 

\noindent \textBF{Task free continual learning.} Recent works have driven the attention to a more challenging scenario where task boundaries are unknown. Most approaches would focus on the sample selection approach that stores certain samples into a buffer to train the model. This approach was firstly investigated in \cite{taskFree_CL} for training a classifier under TFCL and for training both classifiers and VAEs \cite{OnlineContinualLearning} using a new retrieving mechanism selecting called the Maximal Interfered Retrieval (MIR). The Gradient Sample Selection (GSS) \cite{GradientLifelong} formulates the sample selection as a constrained optimization reduction. More recently, a Learner-Evaluator framework, called the Continual Prototype Evolution (CoPE) \cite{ContinualPrototype} stores the same number of samples for each class in the memory to enforce the balance replay. Different from these approaches, the proposed OCM does not require any supervised signals for the sample selection in both supervised and unsupervised learning.

Another approach for TFCL is based on the dynamic expansion mechanism \cite{NeuralDirichelt}, called the Continual Neural Dirichlet Process Mixture (CN-DPM), which introduces Dirichlet processes for the expansion of VAE components. This expansion mechanism was combined with the generative replay into the Continual Unsupervised Representation Learning (CURL) \cite{LifelongUnsupervisedVAE}, for learning the shared and task-specific representations, befitting on the clustering task. 

\noindent \textBF{Optimal Transport (OT).} The OT aims to search for a minimal effort solution to transfer the mass from one distribution to another. OT has been recently applied in the domain adaptation problems \cite{OptimalTransport,UnbalancedOT} and was also used in auto-encoders to provide a flexible training loss for the VAE \cite{WVAE}. However, these models require to fully access all samples at all times, and are failing to capture the underlying data distributions under TFCL. In this paper, we formulate TFCL as the dynamic optimal transport problem which provides a new perspective for the forgetting behaviour of VAEs. To our best knowledge, this paper is the first work to employ OT for forgetting analysis under TFCL.

\section{Preliminary}

In this section, we firstly introduce the background of VAEs. Then we explain how TFCL can be seen as a dynamic optimal transport problem. 

\subsection{The Variational Autoencoder}

The VAE \cite{VAE} aims to jointly optimize the observed variable ${\bf x}$ and their corresponding encoded latent variables ${\bf z}$ within an unified optimization framework by maximizing the marginal log-likelihood $\log p_{\theta} ({\bf x}) = \int p_{\theta} ({\bf x} \,|\, {\bf z}) p({\bf z}) \, \mathrm{d}{\bf z}$. This integral involves the Normal prior distribution $p({\bf z})$, which is intractable to optimize since it requires access to all ${\bf z}$. The VAE maximizes the Evidence Lower Bound (ELBO) on $\log p_{\theta} ({\bf x})$, while the distribution $p_\theta({\bf z} \,|\, {\bf x})$ is approximated by a variational distribution $q_\omega ({\bf z} \,|\, {\bf x})$~:
\begin{equation}
\begin{aligned}
 {\mathcal{L}_{ELBO}}({\bf{x}};\theta ,{\omega} )  : = & {{\mathbb{E}}_{z \sim q_\omega ({\bf z} \,|\, {\bf x})}}\left[ {\log {p_\theta }({\bf{x}} \,|\, {\bf{z}})} \right] - {{KL}}\left[ {q_\omega ({\bf z} \,|\, {\bf x}) \,||\, p({\bf{z}})} \right]\,,
\label{VAE_loss}
\end{aligned}
\end{equation}
where $p_\theta({\bf x} \,|\,{\bf z} )$ is the decoder parameterized by $\theta$ and ${\mathcal{L}_{ELBO}}({\bf{x}};\theta ,{\omega} )$ is a lower bound to $\log p_\theta({\bf x})$. $KL(\cdot)$ represents the Kullback–Leibler (KL) divergence. Eq.~\eqref{VAE_loss} can be further extended when considering multiple samples, as the Importance Weighted Autoencoder (IWVAE) \cite{IWVAE}~:
\begin{equation}
\begin{aligned}
{{\cal L}^m_{IW}}({\bf{x}};\theta ,\omega ): ={\mathbb{E}}_{{{z_1},\cdots,{z_m} \sim {q_\omega }({\bf{z}} \,|\,  {\bf{x}})}} \Bigg[ {\log \frac{1}{m}\sum\limits_{i = 1}^m w_i } \Bigg]
\label{IWVAE_loss}\,,
\end{aligned}
\end{equation}
\noindent where $w_i = {{p_\theta }({\bf{x}}, {\bf{z}}_i)} / {{q_\omega }({\bf{z}}_i \,|\, {\bf{x}})}$ and $m$ is the number of importance samples. Since we have ${\mathcal{L}}^m_{IW}({\bf x}; {\theta},\omega) > {\mathcal{L}}_{ELBO}({\bf x}; {\theta},\omega)$ for $m > 1$ \cite{IWVAE}, Eq.~\eqref{IWVAE_loss} can be used as the estimator for the data likelihood \cite{ImportanceWeightedHierarchical}. 

\subsection{Formulate TFCL as a dynamic OT problem}

\noindent \textbf{Learning setting.} Let ${\mathcal{D}}^{S}$ be a training set over the image space $\mathcal{X} \in \mathbb{R}^d$ with $d$ dimensions, we assume that there are $N$ training steps $\{t_1,\cdots,t_N \}$, for the part-by-part learning of ${\mathcal{D}}^S$, defined as $\mathcal{D}^S = \bigcup_{i=1}^{t_N} {\bf X}^i_b $, where ${\bf X}^i_b  \cap {\bf X}^j_b = \varnothing$ for $i \ne j$.
In each training step $t_i$, a model only observes a small batch of images ${\bf X}^i_b$ drawn from ${\mathcal{D}}^{S}$, without accessing all the prior batches $\{ {\bf X}^{1}_b,\cdots,{\bf X}^{i-1}_b\}$. Once all training steps are finished, we evaluate the model on a testing dataset ${\mathcal{D}}^T$ by using two main criteria (negative log-likelihood estimation and reconstruction quality). In the following, we introduce several definitions and notations.

\begin{definition} (\textBF{Memory.}) Let $\mathcal{M}_{i}$ represent a memory data buffer updated at the step $t_i$ and ${\mathbb{P}}_{m_i}$ represent the probabilistic representation of the samples drawn from $\mathcal{M}_{i}$. Let ${\mathbb{P}}_{\bf x}$ represent the probabilistic measure defined by the samples drawn from $\mathcal{D}^S$.
\end{definition}

\begin{definition} 
(\textBF{Model.}) Let $h^i$ be a VAE model trained on $\mathcal{M}_i$ at $t_i$. Let $\mathbb{P}_{\bf z}$ be a prior distribution (Normal distribution) on the latent variable space $\mathcal{Z}$. 
\end{definition}
\begin{definition} 
\label{definition3} (\textBF{Decoder.})
Let ${\rm G}_i \colon {\mathcal{Z}} \to {\mathcal{X}}$ be a generator (decoder in the $h^i$ model trained at $t_i$). ${\rm G}_i({\bf z}) = p_\theta({\bf x} \,|\, {\bf z})$ in $h^i$ is implemented as the Gaussian decoder ${\mathcal{N}}({\rm G}^\star_i({\bf z}), {\sigma ^2} {\bf I}_d )$, where ${\rm G}^\star_i$ is a deterministic generator, $\sigma > 0$ represents a small random variation for ensuring randomness, and ${\bf I}_d$ is the unit vector of dimension $d$. Let $\mathbb{P}_{{\rm G}_i}$ represent the probabilistic measure formed by samples drawn through the sampling process, ${\bf x} \sim p_\theta({\bf x} \,|\, {\bf z}), {\bf z} \sim {\mathbb{P}}_{\bf z}$ of $h^i$.
\end{definition}

In the generative modelling, we usually consider two probabilistic measures ${\mathbb{P}}_{\bf x}$ and ${\mathbb{P}}_{{\rm G}_i}$ over two distinct spaces, denoted as $\Omega_{\bf x}$ and $\Omega_{{\rm G}_i} $, respectively. Let ${\bf T} \colon \Omega _{{{\rm G}_i}} \to {\Omega _{{\bf x}}}$ be a transport map if satisfying ${\bf T} \# {\mathbb{P}}_{{\rm G}_i} = {\mathbb{P}}_{\bf x}$ that transforms $\mathbb{P}_{{\rm G}_i}$ into $\mathbb{P}_{\bf x}$. For a given arbitrary measurable cost function $\mathcal{L}$, the optimal transportation problem can be defined by Monge's formulation, expressed by~:
\begin{equation}
\begin{aligned}
{{\bf{T}}^*} =& \mathop {\arg \min }\limits_{\bf{T}} \int_{{\Omega _{{{\rm G}_i}}}} {{\mathcal{L}} ({\bf{x}},{\bf{T}}({\bf{x}})) \, \mathrm{d} {\mathbb{P}}_{{\rm G}_i}({\bf{x}})}  \,, s.t.{\bf T} \# {\mathbb{P}}_{{\rm G}_i} = {\mathbb{P}}_{\bf x}\,.
\label{optimalTransport1}
\end{aligned}
\end{equation}
According to the optimal transport theory \cite{OptimalTransport}, the above problem is solved by the Kantorovitch formulation \cite{OptimalTransportBook}~:
\begin{equation}
\begin{aligned}
&{{\mathop{\rm W}\nolimits}^\star _{\mathcal{L}}}({\mathbb{P}}_{\bf x},{\mathbb{P}}_{{\rm G}_i} ) = \mathop {\inf }\nolimits_{ {\mathbb{P}}_{{\bf x} \times {{\rm G}_i}} }  {{{ {{{\mathbb{E}}_{({\bf x}^r,{\bf x}^g ) \sim  {\mathbb{P}}_{{\bf x} \times {{\rm G}_i}} }}[ {{\mathcal{L}}}{{({{\bf{x}}^r},{\bf x}^g )}}} }}} ]\,,
\label{WDistance2}
\end{aligned}
\end{equation}
where ${\mathbb{P}}_{{\bf x} \times {{\rm G}_i}}$ represents the set of all probabilistic couplings on $ {\Omega _{\bf x}} \times {\Omega _{{{\rm G}_i}}}$ with marginals $\mathbb{P}_{\bf x}$ and $\mathbb{P}_{{\rm G}_i}$. ${\bf X}^r$ and ${\bf X}^g$ are the samples drawn from ${\mathbb{P}}_{{\bf x} \times {{\rm G}_i}}$. Different from the traditional OT problem, ${{\mathop{\rm W}\nolimits} _{\mathcal{L}}}({\mathbb{P}}_{\bf x},{\mathbb{P}}_{{\rm G}_i} )$ would be changed over time (when $i$ increases) because the model is trained on the dynamically evolved memory $\mathcal{M}_i$. We call Eq.~\eqref{WDistance2} as the dynamic OT problem where the optimal solution is evolved each training time $t_i$. Eq.~\eqref{WDistance2} has an upper bound when ${\rm G}_i$ is the Gaussian decoder \cite{WVAE2,WVAE}~: 
\begin{equation}
\begin{aligned}
&{{\mathop{\rm W}\nolimits}^\star _{\mathcal{L}}}({\mathbb{P}}_{\bf x},{\mathbb{P}}_{{\rm G}_i} ) \le  \mathop {\inf }\nolimits_{q_\omega ({\bf z})}  {{{ {{{\mathbb{E}}_{ {\mathbb{P}}_{\bf x} }} {\mathbb{E}}_{q_\omega({\bf z} \,|\, {\bf x})}[ {{\mathcal{L}}}{{({{\bf{x}}}, {\rm G}_i({\bf z}) )}}} }}}]\,,
\label{WDistance3}
\end{aligned}
\end{equation}
where $q_\omega({\bf z})$ is the marginal distribution of $q_\omega({\bf z} \,|\, {\bf x})$ satisfying $q_\omega ({\bf z}) = p(\bf z)$. We implement $\mathcal{L}({\bf x} ,{\rm G}_i({\bf z}) ) =\left\| { {\bf x} - {\rm G}_i({\bf z} }) \right\|^2 $ as the squared Euclidean cost function in which ${\rm W}_{\mathcal{L}}(\cdot)$ is the squared 2-Wasserstein distance \cite{WVAE2}.

\section{Theoretical framework}
\label{sec:theory}
ELBO is an important indicator of the VAE's performance and is used as its main optimization function \cite{VAE_symmetric}. In the following, we provide a new perspective for analyzing the forgetfulness behaviour of VAEs during the continuous learning of several batches of data by formulating the ELBO's variation as a learning and forgetting process. The code and Supplemental Materials (SM) are available at \url{https://github.com/dtuzi123/OVAE}.

\subsection{Analysis of forgetting in a single model}

Firstly, we derive an upper bound to ELBO of the target domain ${\mathbb{P}}_{\bf x}$, based on the dynamic OT problem (Eq.~\eqref{WDistance3}).

\begin{theorem}
\label{theorem1} 
For a VAE model $h^i$ trained at $t_i$, where $p_\theta({\bf x \,|\, {\bf z}}) = {\mathcal{N}}({\rm G}_i({\bf z}), {\sigma ^2} {\bf I}_d )$ is the Gaussian decoder and $\sigma = 1/\sqrt 2 $, we have~:
\begin{equation}
\begin{aligned}
\mathop {\inf }\nolimits_{q_\omega ({\bf z}) = p(\bf z)}  {{{ {{{\mathbb{E}}_{ {\mathbb{P}}_{\bf x} }} }}}} [ {\mathcal{L}}_{ELBO}({\bf x} ; \theta,\omega )  ] 
 &\le - \frac{1}{2}\log \pi - {\rm W}^\star_{\mathcal{L}}({\mathbb{P}}_{\bf x},{\mathbb{P}}_{{\rm G}_i})
 \,,
 \label{theorem1_eq1}
\end{aligned}
\end{equation}
\end{theorem}
The detailed proof is provided in Appendix-A from Supplemental Materials (SM). Based on the results from Theorem~\ref{theorem1}, we derive a bound that explains the forgetting process of VAEs.

\begin{theorem} 
\label{theorem2}
Let $\mathbb{P}_{m_i}$ and $\mathbb{P}_{\bf x}$ be the source and target domains. From Eq.~\eqref{theorem1_eq1}, we derive the bound on the ELBO between $\mathbb{P}_{m_i}$ and $\mathbb{P}_{\bf x}$ at the training step $t_i$~:
\begin{equation}
\begin{aligned}
  {\mathbb{E}}_{ {\mathbb{P}}_{\bf x} } [ {\mathcal{L}}_{ELBO}({\bf x} ; \theta,\omega )  ]
 &\le 
 {{{ {{{\mathbb{E}}_{ {\mathbb{P}}_{m_i} }} [ {{\mathcal{L}}}_{ELBO}{{({{\bf{x;\theta,\omega }}} )}}} }}}]
 \\&+
  2{\rm{W}}_{\cal L}^ \star ({{\mathbb{P}}_{{m_i}}},{{\mathbb{P}}_{{{\rm{G}}_i}}})
- {\rm W}^\star_{\mathcal{L}}({\mathbb{P}}_{\bf x},{\mathbb{P}}_{m_i}) +{\widetilde {\rm F}}({\mathbb{P}}_{{\rm G}_i},{\mathbb{P}}_{m_i})
 \,, 
 \label{theorem2_eq1}
\end{aligned}
\end{equation}
\noindent where ${\widetilde {\rm F}}({\mathbb{P}}_{{\rm G}_i},{\mathbb{P}}_{m_i})$ is expressed as~:
\begin{equation}
\begin{aligned}
 {\widetilde {\rm F}}({\mathbb{P}}_{{\rm G}_i},{\mathbb{P}}_{m_i}) &= {{{\mathbb{E}}_{ {\mathbb{P}}_{m_i} }} [  D_{KL}(q_{\omega }({\bf z}\,|\,{\bf x} ) \,||\,p({\bf z})) 
 } ]  \\&+ \Big|
 {{\mathbb{E}} _{{{\mathbb{P}}_{{m_i}}}}}{{\mathbb{E}}_{{q_\omega }({\bf{z}} \,|\,  {\bf{x}})}}[ - {\cal L}({\bf{x}},{{\rm{G}}_i}({\bf{z}}))] -  {\rm{W}}_{\cal L}^ \star ({{\mathbb{P}}_{{m_i}}},{{\mathbb{P}}_{{{\rm{G}}_i}}})  
\Big| 
 \,.
\end{aligned}
\end{equation}
\end{theorem}

\noindent \textBF{Remark.} The detailed proof is provided in Appendix-B from SM. We have several observations from Theorem~\ref{theorem2}~:
\begin{inparaenum}[1)]
\item Improving the performance on the source domain (ELBO on $\mathbb{P}_{m_i}$) would not lead to increasing ELBO on the target domain $\mathbb{P}_{\bf x}$ because the right hand side (RHS) of Eq.~\eqref{theorem2_eq1} involves the negative term, $-{\rm W}^\star_{\mathcal{L}}({\mathbb{P}}_{\bf x},{\mathbb{P}}_{m_i})$.
\item Since RHS of Eq.~\eqref{theorem2_eq1} is upper bounded to ELBO on ${\mathbb{P}}_{\bf x}$, a large ${\rm W}^\star_{\mathcal{L}}({\mathbb{P}}_{\bf x},{\mathbb{P}}_{m_i})$ decreases RHS of Eq.~\eqref{theorem2_eq1} and therefore leads to the degenerated performance, measured by ELBO, on $\mathbb{P}_{\bf x}$, corresponding to forgetting the knowledge at the training step $t_i$. This is usually caused by the memory $\mathcal{M}_i$ that fails to capture all information of $\mathbb{P}_{\bf x}$ during the initial training process (when $i$ is small) or after the training ($i=t_N$).
\end{inparaenum}

\noindent \textBF{The effect of the memory diversity.} In practice, $\mathbb{P}_{\bf x}$ is divided into several separate distributions (target domains) $\{ {\mathbb{P}}_{{\bf x}^1},\cdots,{\mathbb{P}}_{{\bf x}^n} \}$ where each $\mathbb{P}_{{\bf x}^j}$ is the characteristic distribution of a data category. Under this setting we analyze the forgetting behaviour in the class-incremental scenario.
\begin{lemma}
\label{lemma1}
Let $\{ {\mathbb{P}}_{{\bf x}^1},\cdots,{\mathbb{P}}_{{\bf x}^n} \}$ and $\mathbb{P}_{m_i}$ be the target domains and source domain, respectively. The bound on ELBO between the source and target domain is derived as~:
\begin{equation}
\begin{aligned}
&\sum\nolimits_{j = 1}^n \, { {{{\mathbb{E}}_{{{\mathbb{P}}_{{{\bf{x}}^j}}}}}[ {{\cal L}_{ELBO}}({\bf{x}};\theta ,\omega )]} }  \le \sum\nolimits_{j = 1}^n \Big\{
2{\rm{W}}_{\cal L}^ \star ({{\mathbb{P}}_{{m_i}}},{{\mathbb{P}}_{{{\rm{G}}_i}}})
\\&+
 {{{ {{{\mathbb{E}}_{ {\mathbb{P}}_{m_i} }} [ {{\mathcal{L}}}_{ELBO}{{({{\bf{x;\theta,\omega }}} )}}} }}}]
- {\rm W}^\star_{\mathcal{L}}({\mathbb{P}}_{{\bf x}^j},{\mathbb{P}}_{m_i})
 \Big\}  + n{\widetilde {\rm F}}({\mathbb{P}}_{{\rm G}_i},{\mathbb{P}}_{m_i})\,.
 \label{lemma1_eq1}
\end{aligned}
\end{equation}
\end{lemma}

\noindent \textBF{Proof.} We sum up the bounds between $\mathbb{P}_{{\bf x}^j}$ and $\mathbb{P}_{m_i}$, where $j=1,\cdots,n$ and prove Lemma~\ref{lemma1}.

\noindent \textBF{Remark.} We have several observations from Lemma~\ref{lemma1}~:
\begin{inparaenum}[1)]
\item To maximize ELBO on target domains $\{ {\mathbb{P}}_{{\bf x}^1},\cdots,{\mathbb{P}}_{{\bf x}^n} \}$, ${\rm W}^\star_{\mathcal{L}}({\mathbb{P}}_{{\bf x}^j} ,{\mathbb{P}}_{m_i} ), j=1,\cdots,n$ must be minimized, corresponding to the diverse samples replayed from $\mathbb{P}_{m_i}$.
\item We also provide new insights into the backward transfer \cite{GradientEpisodic} by using Eq.~\eqref{lemma1_eq1}. When a memory $\mathcal{M}_i$ prefers to store samples from a few recent data distributions $\{ {\mathbb{P}}_{{\bf x}^{n-1}},{\mathbb{P}}_{{\bf x}^{n}} \}$, the model would lead to negative backward transfer on past target domains $\{ {\mathbb{P}}_{{\bf x}^{1}},\cdots,{\mathbb{P}}_{{\bf x}^{n-2}} \}$. Data diversity in memory can relieve this negative effect.  
\end{inparaenum}

\subsection{Forgetting analysis of the expanding VAE mixture model}

In this section, we extend the forgetting analysis from a single VAE model to the Dynamic Expansion Model (DEM).
\begin{definition} 
\label{defintion4}
Let ${\bf H} = \{h_1,\cdots,h_k \}$ be a dynamic expansion model trained at $t_i$, which has built $k$ components during the learning, where each $h_i$ is a VAE model. Let ${\bf q}=\{q_1,\cdots,q_k \}$ represent the training steps that each component converged on. For instance, $h_i$ converged on $\mathcal{M}_{q_i}$ at $t_{q_i}$, is not updated in the following training steps. Then  $\, \mathbb{P}_{{\rm G}_{q_i}}$ and  $\, \mathbb{P}_{m_{q_i}}$ represent the generator distribution and the distribution of samples drawn from $\mathcal{M}_{q_i}$.
\end{definition}

\begin{lemma} 
\label{lemma2}
Let $\{ {\mathbb{P}}_{{\bf x}^1},\cdots,{\mathbb{P}}_{{\bf x}^n}\}$ be a set of $n$ target domains. From Definition~\ref{defintion4}, the bound on the ELBO for the dynamic expansion model is derived as~:
\begin{equation}
\begin{aligned}
&\sum\nolimits_{j = 1}^n { {{{\mathbb{E}}_{{{\mathbb{P}}_{{{\bf x}^j}}}}}[{{\cal L}_{ELBO}}({\bf{x}};\theta ,\omega )]} }  \le \sum\nolimits_{i = 1}^n 
{\rm F}^\star({\mathbb{P}}_{{\bf x}^i}) \,,
 \label{lemma2_eq1}
\end{aligned}
\end{equation}
where ${\rm F}^\star({\mathbb{P}}_{{\bf x}^i})$ is the selection function, defined as~:
\begin{equation}
\begin{aligned}
{\rm F}^\star({\mathbb{P}}_{{\bf x}^i}) &= \mathop {\max }\limits_{j = 1,\cdots,k} \Big\{
 {{{ {{{\mathbb{E}}_{ {\mathbb{P}}_{m_{q_j}} }} [ {{\mathcal{L}}}_{ELBO}{{({{\bf{x;\theta,\omega }}} )}}} }}}] 
\\&+
2{\rm{W}}_{\cal L}^ \star ({{\mathbb{P}}_{{m_{q_j}}}},{{\mathbb{P}}_{{{\rm{G}}_{q_j}}}}) - {\rm W}^\star_{\mathcal{L}}({\mathbb{P}}_{{\bf x}^i},{\mathbb{P}}_{m_{q_j}})
+{\widetilde {\rm F}}({\mathbb{P}}_{{\rm G}_{q_j}},{\mathbb{P}}_{m_{q_j}})
 \Big\}\,.
\label{selectFun}
\end{aligned}
\end{equation}
\end{lemma}

The proof is provided in Appendix-C from SM. To compare with a single model (Lemma~\ref{lemma1}), DEM would provide a maximum upper bound to the Left Hand Side (LHS) of Eq.~\eqref{lemma2_eq1} due to the selection process, Eq.~\eqref{selectFun}. Additionally, DEM can relieve the negative backward transfer by preserving prior knowledge into the frozen components.

\subsection{Mixture expansion with the task information}
\label{Expand}

Although the proposed theoretical framework is only used for TFCL, it can be extended for the case where task labels are known. We also apply the proposed theoretical framework for analyzing the forgetting behaviour of existing approaches (See details in Appendix-F from SM).

\begin{definition} (\textBF{ Learning setting.}) Let $\mathcal{T} = \{ \mathcal{T}_1,\cdots,\mathcal{T}_c \}$ represent a set of task labels where $c$ is the number of tasks and we consider that each $i$-th task is associated with a testing dataset $\mathcal{D}_i^T$ and a training dataset $\mathcal{D}_i^S$. Let $\mathbb{P}_{{\bf x}^i}$ and $\mathbb{P}_{{\widehat{\bf x}}^i}$ represent the empirical distributions for $\mathcal{D}_i^S$ and $\mathcal{D}_i^T$, respectively. Since the task label is given, a mixture model starts to learn the first task and then either builds a new component or selects an existing component to learn a new task after the task switch. When a certain component is selected to learn a new task, the Generative Replay Mechanism (GRM) is used to relieve forgetting.
\end{definition}

\begin{definition} (\textBF{Generative replay.}) Let $\mathbb{P}^j_{{\widetilde{\bf x}}}$ represent the distribution of samples drawn from the generating process of $h_j$. Let $f_t \colon {\mathcal{X}} \to {\mathcal{T}}$ be the true labelling function that returns the task label for the data sample. If the $i$-th task is trained by $h_j$, let $\mathbb{P}_{{\widetilde{\bf x}}^{(i,m)}}$ be the distribution of samples drawn from the process ${\bf x} \sim \mathbb{P}^j_{{\widetilde{\bf x}}}$ if ${f_t({\bf x}) = i }$, where $m$ represents that $\mathbb{P}_{\widetilde{\bf x}^{(i,0)}}$ is evolved to $\mathbb{P}_{\widetilde{\bf x}^{(i,m)}}$ through $m$ generative replay processes \cite{LifelongVAEGAN}. Let $\mathbb{P}_{\widetilde{\bf x}^{(i,0)}}$ and $\mathbb{P}_{\widetilde{\bf x}^{(i,-1)}}$ represent ${\mathbb{P}}_{{\bf x}^i}$ and ${\mathbb{P}}_{{ \widehat{\bf x}}^i}$ for simplicity.
\end{definition}

\begin{theorem} 
\label{theorem3}
Let $\mathcal{A}=\{ a_1,\cdots,a_n \}$ be a set where each $a_i$ represents the index of the component that has trained only once. Let ${\widetilde{{\mathcal{A}}}} = \{ {{\widetilde a}_1},\cdots,{{\widetilde a}_{n}} \}$ be a set of task labels where each ${\widetilde a}_i$ represents the index of the task learned by the $a_i$-th component. Let $\mathcal{B} = \{b_1,\cdots,b_{k-n} \}$ be a set where each $b_i$ represents the index of the component that is trained more than once. Let ${\widetilde b}_i = \{ {\widetilde b}_i^1,\cdots, {\widetilde b}_i^m \}$ be a set of task labels for the $b_i$-th component. Let ${ c}_i^j$ represent the number of generative replay processes for the ${\widetilde b}_{i}^j$-th task, achieved by the $b_i$-th component. Let $\mathbb{P}_{{\rm G}^i}$ represent the generator distribution of the $i$-th component. We derive the bound for a mixture model with $k$ components trained on $c$ tasks as~:
\begin{equation}
\begin{aligned}
&\sum\limits_{i = 1}^{|{\mathcal{A}}|} \Big\{
  {\mathbb{E}}_{ {\mathbb{P}}_{{\widehat{\bf x}}^{{\widetilde{a}}_i }} } [ {\mathcal{L}}_{ELBO}({\bf x} ; \theta,\omega )  ]
\Big\} + \sum\limits_{i = 1}^{|{\mathcal{B}}|} \Big\{ {\sum\limits_{q = 1}^{| {\widetilde b}_i |}  } \big\{
  {\mathbb{E}}_{ {\mathbb{P}}_{{\widehat{\bf x}}^{{\widetilde{b}}_i^q }} } [ {\mathcal{L}}_{ELBO}({\bf x} ; \theta,\omega )  ] \big\} \Big\}
 \le \mathcal{R}_{S} + \mathcal{R}_{M}
 \label{theorem3_eq}
\end{aligned}
\end{equation}
where $|\cdot|$ denotes the cardinal of a set. $\mathcal{R}_{S}$ is estimated by components that are trained only once, defined as~:
\begin{equation}
\begin{aligned}
{\mathcal{R}}_{S} &=  \sum\nolimits_{i = 1}^{|{\mathcal{A}}|} \Big\{
2{\rm{W}}_{\cal L}^ \star ({\mathbb{P}}_{{\bf x}^{{\widetilde{a}}_i }} ,{{\mathbb{P}}_{{{\rm{G}}^{a_i}}}}) +
 +{\widetilde {\rm F}}({\mathbb{P}}_{{\rm G}^{a_i}}
{\mathbb{P}}_{{{\bf x}}^{{\widetilde{a}}_i }})
 \\&+
  {{{ {{{\mathbb{E}}_{ {\mathbb{P}}_{{\bf x}^{{\widetilde{a}}_i }} }} [
 {{\mathcal{L}}}_{ELBO}{{({{\bf{x;\theta,\omega }}} )}}} }}}]
- {\rm W}^\star_{\mathcal{L}}(
{\mathbb{P}}_{{\widehat{\bf x}}^{{\widetilde{a}}_i }}
,{\mathbb{P}}_{{{\bf x}}^{{\widetilde{a}}_i }})  \Big\}.
\end{aligned}
\end{equation}
$\mathcal{R}_{M}$ is estimated by components that are trained on more than one task, as~:
\begin{align}
   {\mathcal{R}}_{M} &=  \sum\nolimits_{i = 1}^{|{\mathcal{B}}|} \Big\{ {\sum\nolimits_{q = 1}^{| {\widetilde b}_i |}  } \Big\{ 
  {{{ {{{\mathbb{E}}_{ {\mathbb{P}}_{{\widetilde{\bf x}}^{({\widetilde{b}}_i^q, c_i^q ) }} }} [ {{\mathcal{L}}}_{ELBO}{{({{\bf{x;\theta,\omega }}} )}}} }}}]
 +
\sum\nolimits_{s = 0}^{c_i^q} \big\{
  2{\rm{W}}_{\cal L}^ \star ({\mathbb{P}}_{{\widetilde{\bf x}}^{({\widetilde{b}}_i^q,s ) }} ,{{\mathbb{P}}_{{{\rm{G}}^{b_i}}}}) \notag
\\&+{\widetilde {\rm F}}({\mathbb{P}}_{{\rm G}^{b_i}},
{\mathbb{P}}_{{\widetilde{\bf x}}^{({\widetilde{b}}_i^q,s) }}
)
- 
{\rm W}^\star_{\mathcal{L}}(
{\mathbb{P}}_{{\widehat{\bf x}}^{({\widetilde{b}}_i^q,s-1 )}}
,{\mathbb{P}}_{{\widetilde{\bf x}}^{({\widetilde{b}}_i^q,s ) }})  \big\} \Big\} \Big\}.
\label{RM_eq} 
\end{align}
\end{theorem}
\noindent \textBF{Remark.} The detailed proof is provided in Appendix-D from SM. Theorem~\ref{theorem3} has the following observations~:
\begin{inparaenum}[1)]
\item If the number of components $k$ is equal to the number of tasks, then $\mathcal{R}_M = 0$ and there is no forgetting. When the number of components decreases, forgetting happens because the last term in the RHS of Eq.~\eqref{RM_eq} is increased, leading to a decrease in the RHS of Eq.~\eqref{theorem3_eq} (corresponding to the decrease of ELBO on all target domains). 
\item If $k = 1$, then $\mathcal{R}_S$ is about only the last task, then $\mathcal{R}_M$ is increased significantly since the accumulated errors $\sum\nolimits_{s = 0}^{c_i^q} \{ {\rm W}^\star_{\mathcal{L}}(
{\mathbb{P}}_{{\widehat{\bf x}}^{({\widetilde{b}}_i^q,s-1 )}}
,{\mathbb{P}}_{{\widetilde{\bf x}}^{({\widetilde{b}}_i^q,s ) }}) \}$ in Eq.~\eqref{RM_eq} increases. Learning early tasks would lead to more forgetting than when learning the recent tasks for $k=1$ because early tasks would have more accumulated errors ($c_i^q$ in $\mathcal{R}_M$ is large as $i$ increases (See Appendix-D from SM)).
\end{inparaenum}

\section{Methodology}
\label{sec:methodology}

Previous approaches have proposed to learn a diverse memory according to the category information. However, these approaches do not provide a theoretical guarantee for the accumulated memory's diversity. To our best knowledge, this paper is the first to provide a theoretical forgetting analysis and guarantees for existing TFCL models (See details in Appendix-F of SM). Additionally, the proposed theoretical framework demonstrates that the diversity of memory content can be achieved without knowing the category information (Lemma~\ref{lemma1}). Based on the conclusion of Lemma~\ref{lemma1}, we introduce a new memory approach which consists of three modules: LTM, STM and the Learner. The proposed approach does not require any task information or supervised signals for unsupervised learning. Firstly, we introduce the proposed OCM with the Learner implemented as a single VAE, and then we extend this into a dynamic expansion mechanism.

\subsection{Online Cooperative Memorization (OCM)}
\label{sec:onlineMemory}

\noindent \textBF{Notations.}  Let $\mathcal{M}^l_i=\{ 
{\bf x}^l_{i,j}\}_{j=1}^{n^l_i}$ and $\mathcal{M}^e_i=\{ 
{\bf x}^e_{i,u} \}_{u=1}^{n_i^e}$ represent the samples stored in the LTM and STM, respectively, at the training step $t_i$ while $n_i^l$ and $n_i^e$ represent the number of samples. Let $\mathcal{M}_{Max}^e $ represent the maximum number of samples which can be stored in $\mathcal{M}^e_i$. 

\begin{figure}[t]
\centering
		\includegraphics[scale=0.195]{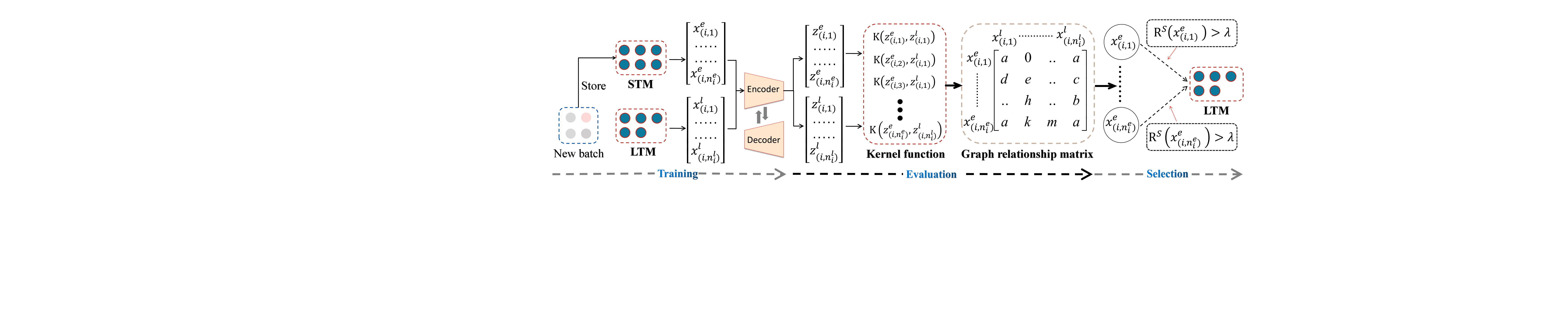}
	\caption{
	The training of OCM consists of three stages~: \textBF{(Learning.)} STM continually stores recent samples while the model is trained to adapt both LTM and STM; If STM is full, we perform the evaluation and selection stages, otherwise, we continually perform the learning stage. \textBF{(Evaluation.)} We obtain the feature vectors $\{ {\bf z}^e_{(i,1)},\cdots,{\bf z}^l_{(i,n^l_i)} \}$ from inputs $\{ {\bf x}^e_{(i,1)},\cdots,{\bf x}^l_{(i,n^l_i)} \}$ by using a VAE encoder, which is used for the evaluation of the sample similarity using the kernel from Eq.~\eqref{kernelFun}. This similarity information is preserved in the graph relationship matrix ${\bf S}_i$. \textBF{(Selection.)} We transfer the samples from STM to LTM using the proposed criterion Eq.~\eqref{criterion} by means of ${\bf S}_i$ from \eqref{kernelFun2}.
	}
	\label{DualMemory}
\end{figure}

The training procedure, presented in Fig.~\ref{DualMemory}, consists of three main stages, as described in the following.

\noindent \textBF{Stage 1 : Learning.} At the training step $t_i$, STM stores a new batch of samples ${\bf X}_i^b$ into $\mathcal{M}^e_i$, while the model, consisting of a single VAE, is trained to update both $\mathcal{M}^e_i$ and $\mathcal{M}^l_i$ using Eq.~\eqref{VAE_loss}. Once the training is finished, we perform the next step.

\noindent \textbf{Stage 2: Evaluation.} 
We perform this step if and only if $n^e_i \ge \mathcal{M}_{Max}^e$ in order to reduce the computational cost. The main goal of this stage is to evaluate the correlation between stored samples from STM and LTM. Firstly, we treat each stored sample as a node and introduce a graph relationship matrix ${\bf S}_i \in \mathbb{R}^{n_i^e \times n_i^l}$, whose elements ${\bf S}_i(j,u)$ represent the correlation between two samples ${\bf x}^e_{i,j}$ and ${\bf x}^l_{i,u}$, from STM and LTM respectively. Directly evaluating each ${\bf S}_i(j,u)$ in the high-dimensional data space is intractable since it would require overloaded computations \cite{KLForMixture} and auxiliary training \cite{MutualInformation,WGAN_Transport}. Since the model has been trained on both past samples from LTM and the current samples from STM, it can be used as a discriminator. We then evaluate the distance between two samples based on the perceptual feature space of the learned model by using the Radial Basis Function (RBF) kernel~:
\begin{equation}
\begin{aligned}
{\mathop{\rm K}\nolimits} ( {\bf x}^e_{i,j},{\bf x}^l_{i,u}) = \exp \left(  - \frac{{\left\| {  {\bf z}^e_{i,j} - {\bf z}^l_{i,u}} \right\|^2 }}{{2{\alpha ^2}}}\right) \,,
\label{kernelFun}
\end{aligned}
\end{equation}
where ${\bf z}^e_{i,j}$ and ${\bf z}^l_{i,u}$ are feature vectors extracted from ${\bf x}^e_{i,j}$ and ${\bf x}^l_{i,u}$ using the feature extractor implemented by the output layer of the encoder $q_\omega ({\bf z} \,|\, {\bf x}) $ of the VAE model, as illustrated in Fig.~\ref{DualMemory}. ${\bf S}_i(j,u) = {\mathop{\rm K}\nolimits} ( {\bf x}^e_{i,j},{\bf x}^l_{i,u})$ and $\left\| \cdot \right\|^2$ is the squared Euclidean distance. $\alpha$ is the scale hyperparameter for the kernel and we set $\alpha = 10$ to ensure that the output of ${\rm K}(\cdot,\cdot)$ is within $[0,1]$. Eq.~\eqref{kernelFun} can be further accelerated by the matrix operation, expressed as~:
\begin{equation}
\begin{aligned}
{{\bf{S}}_i} = {{\mathop{\rm F}\nolimits} _{\exp }}\Big( { - ({\bf{Z}}_i^e{{( - {\bf{Z}}_i^l)}^{\mathop{\rm T}\nolimits} }) \odot ({\bf{Z}}_i^e{{( - {\bf{Z}}_i^l)}^{\mathop{\rm T}\nolimits} })/2{\alpha ^2}} \Big),
\label{kernelFun2}
\end{aligned}
\end{equation}
where ${\bf Z}^e_i \in {\mathbb{R}}^{n_i^e\times d_z}$ and ${\bf Z}^l_i \in {\mathbb{R}}^{n_i^l\times d_z}$ are the feature matrices corresponding to $\mathcal{M}^e_i$ and $\mathcal{M}^l_i$, where each row is a feature vector of dimension $d_z$. $(\cdot)^{\rm T}$ and $ \odot $ are the transpose operation and Hadamard product, respectively. ${\rm F}_{exp}(\cdot)$ is the exponential function for each element in a matrix.

\renewcommand\arraystretch{1.1}
\begin{table}[t]
    \centering
    \scriptsize
		\caption{The estimation of log-likelihood on all testing samples by using the IWVAE bound with 1000 importance samples.}
\setlength{\tabcolsep}{1.3mm}{
\begin{tabular}{l c c c c c c c ccc ccc c}
\toprule 
& \multicolumn{3}{c}{\textbf{Split MNIST}}&
\multicolumn{3}{c}{\textbf{Split Fashion}}& 
\multicolumn{3}{c}{\textbf{Split MNIST-Fashion}} 
\\ 
\cmidrule(l){2-4}\cmidrule(l){5-7}\cmidrule(l){8-10}
\textbf{Methods}	& Log &Memory &N& Log &Memory &N& Log &Memory &N \\
\midrule 
		VAE-ELBO-Random &-150.79&3.0K&1&-280.54&3.0K&1&-247.46&3.0K&1
		\\
		LIMix \cite{LifelongInfinite}&  -146.23&2.0K&30&-262.52&2.0K&30&-238.63&2.0K&30  \\
	   CNDPM \cite{NeuralDirichelt}& -120.71&2.0K&30 &-257.56&2.0K&30&-236.79  &2.0K &30   \\
		\hline
		\hline
		VAE-ELBO-OCM&-132.07&1.6K&1 &-250.74&1.6K&1&-215.62&2.0K&1   \\
	VAE-IWVAE50-OCM& -127.11&1.6K&1&-247.90&1.6K&1&-224.34&2.0K&1  \\
	Dynamic-ELBO-OCM& \textBF{-115.89}&1.1K&5 &\textBF{ -237.69}&1.3K&10&\textBF{ -187.49}&1.4K&10 \\
\bottomrule 
	\end{tabular}
	\label{tab:logLikelihood}
	}
\end{table}

\begin{figure}[t]
	\centering
	\subfigure[Real testing samples.]{
		\centering
		\includegraphics[scale=0.27]{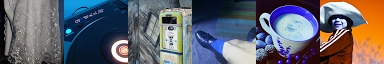}
	}
		\centering
	\subfigure[VAE-ELBO-Random.]{
		\centering
		\includegraphics[scale=0.27]{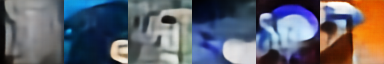}
	}
			\centering
	\subfigure[VAE-ELBO-OCM.]{
		\centering
		\includegraphics[scale=0.27]{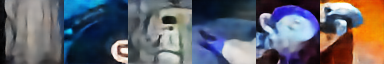}
	}
	\centering
	\caption{Image reconstruction compared to real images.}
	\label{Reco_ImageNet}
\end{figure}

\noindent \textbf{Stage 3: Sample selection.} This stage also require satisfying $N^e_i \ge {\mathcal{M}_{Max}^e}$ to avoid excessive LTM growing. The main goal of this stage is to choose samples that are very different from those already stored in LTM. We achieve this by calculating the average similarity scores using kernels between each candidate sample ${\bf x}^e_{i,j}$ and each sample from LTM using ${\bf S}_i$ from Eq.~\eqref{kernelFun2}~:
\begin{equation}
\begin{aligned}
{{\mathop{\rm R}\nolimits} ^S}({\bf{x}}_{i,j}^e) = \frac{1}{{n_i^l}}\sum\nolimits_{k = 1}^{n_i^l} {{\bf S}_i(j,k)} \,.
\label{similarityScore}
\end{aligned}
\end{equation}
Eq.~\eqref{similarityScore} refers to the distance between ${\bf x}^e_{i,j}$ and all samples contained in the LTM. In order to control the size of LTM, we introduce a threshold $\lambda$ for the sample selection~:
\begin{equation}
\begin{aligned}
{{\mathop{\rm R}\nolimits}^S}({\bf x}_{i,j}^e) > \lambda
 \Rightarrow  \mathcal{M}^l_i  = \mathcal{M}^l_i \cup {\bf x}_{i,j}^e \,.
\label{criterion}
\end{aligned}
\end{equation}
The choice for $\lambda$ influences the diversity and memory size of LTM. Empirically, according to the ablation study in Appendix H.4 from SM, $\lambda \in [0.2,0.5]$ can achieve the best performance resulting in a reasonable LTM size for most datasets. Once the selection is finished, $\mathcal{M}^e_i$ is cleared for storing novel samples during the next training step $t_{i+1}$.

\makeatletter\def\@captype{table}\makeatother
{
\centering
\begin{minipage}{0.46\textwidth}
\centering
    \scriptsize
    \renewcommand\arraystretch{1.28}
	\caption{IS and FID scores under Split CIFAR10.}
\setlength{\tabcolsep}{0.5mm}{\begin{tabular}{@{}l cc cc @{} } 
\toprule 
\textBF{Methods}   &{\em IS}&{\em FID} &Memory&N  \\
\midrule 
	  VAE-ELBO-Random&3.84&116.26&1.0K&1 \\
	  CNDPM \cite{NeuralDirichelt}&4.12&95.23&1.0K&30 \\
	  LIMix \cite{LifelongInfinite}&3.02&156.46&1.0K&30 \\
	 \hline
	 \hline
	   VAE-ELBO-OCM& 4.13&98.76&0.5K&1 \\
	   Dynamic-ELBO-OCM&\textBF{4.16} & \textBF{92.99} & 0.4K&3  \\
\bottomrule 
\end{tabular}
		\label{SplitCIFAR10_TinyImageNet}
}
\end{minipage}
\;\;\;\;\;\;\;\;
\makeatletter\def\@captype{table}\makeatother
\begin{minipage}{.45\textwidth}
	\centering
    \scriptsize
		\caption{The estimation of log-likelihood on “Cross domain”}
\setlength{\tabcolsep}{0.6mm}{
\begin{tabular}{l c c c }
\toprule 
\cmidrule(l){2-4}
\textbf{Methods}	&  Log &Memory &N \\
\midrule 
		VAE-ELBO-Random &-239.71&3.0K&1
		\\
		LIMix \cite{LifelongInfinite} &-226.63&2.0K&30  \\
	   CNDPM \cite{NeuralDirichelt} &-218.15&2.0K&30  \\
		\hline
		\hline
		VAE-ELBO-OCM &-201.31&2.0K&1  \\
	VAE-IWVAE50-OCM&-204.35&2.0K&1  \\
	Dynamic-ELBO-OCM&\textBF{ -177.29}&1.5K&11 \\
\bottomrule 
	\end{tabular}
	\label{tab:logLikelihood2}
	}
\end{minipage}
}

\subsection{Combining OCM with expansion mechanism}
\label{sec:expansion}

According to Lemma~\ref{lemma2} and Section~\ref{Expand}, by dynamically expanding the model with new components would lead to better performance. Moreover, the extension mechanism reduces negative transfer when each component learns different underlying data distributions (see detailed analysis in Appendix-C of SM). This analysis inspires us to implement the extension mechanism from two aspects. First, we introduce an expansion criterion to detect the data distribution shift by comparing the loss value between the previously learned and newly seen samples, which ensures a suitable network architecture. Second, to encourage each component to learn different underlying data distributions, we clear STM and LTM when we dynamically add a new component to the mixture model.

The newly added component can be an independent VAE or one that shares its parameters with existing components. In the following, we describe the latter setting. Let ${\rm f}_{{\omega _s}}^e \colon \mathcal{X} \to \mathcal{Z}'$ and ${\rm f}_{{\omega _i}}^e \colon \mathcal{Z}' \to \mathcal{Z}$ be the shared module and the component-specific module for the encoding process, where $i$ represents the component index and $\mathcal{Z}'$ is the feature space. Similar to the encoding process, we have two modules for the decoding process, ${\rm f}_{{\theta _s}}^d \colon \mathcal{Z} \to \mathcal{X}'$ and ${\rm f}_{{\theta _i}}^d \colon \mathcal{X}'\to \mathcal{X}$, where $\mathcal{X}'$ is the feature space. The encoding and decoding processes for the $i$-th component can be implemented by $q_{\theta_{s,i}} ({\bf z} \,|\, {\bf x}) = {\rm f}_{{\omega _s}}^e \odot  {\rm f}_{{\omega _i}}^e({\bf x})$ and $p_{\theta_{s,i}}({\bf x} \,|\, {\bf z}) =  {\rm f}_{{\theta _s}}^e \odot  {\rm f}_{{\theta _i}}^e({\bf z})$, respectively, where ${\rm f}_{{\omega _s}}^e \odot  {\rm f}_{{\omega _i}}^e \colon {\mathcal{X} \to {\mathcal{Z}}' \to {\mathcal{Z}} }$ is the encoding process. The optimization for the $i$-th component corresponds to maximizing ELBO~:
\begin{equation}
\begin{aligned}
 {\mathcal{L}^i_{ELBO}}({\bf{x}};\theta ,{\omega} ) & : =   {{\mathbb{E}}_{ q_{\omega_{s,i}} ({\bf z} \,|\, {\bf x})}}\left[ {\log {p_{\theta_{s,i}} }({\bf{x}} \,|\, {\bf{z}})} \right] - {{KL}}\left[ {q_{\omega_{s,i}} ({\bf z} \,|\, {\bf x}) \,||\, p({\bf{z}})} \right]
\label{mixture_loss}
\end{aligned}
\end{equation}
where ${\bf z} \sim q_{\omega_{s,i}} ({\bf z} \,|\, {\bf x})$ and the shared modules are only updated by using Eq.~\eqref{mixture_loss} for $i > 1$ in order to avoid forgetting.

\noindent \textbf{Criterion for dynamic expansion.}
When a mixture model has multiple components, we evaluate the sample similarity from Eq.~\eqref{kernelFun} by using an augmented feature extractor that concentrates features from each component. The training process for the new components from the dynamic expansion model is the same as the one described in Section~\ref{sec:onlineMemory} where we incorporate a criterion for the model expansion in \noindent \textBF{Step 3 : (Sample selection)}~:
\begin{equation}
\begin{aligned}
| {\mathop{\rm R}}_i - {\mathop{\rm R}}_{last} |
> \lambda_2\,, {{\rm{R}}_i} =  \frac{1}{{N'}}\sum\nolimits_{j = 1}^{N'} {\Big\{ \frac{1}{K}\sum\nolimits_{c = 1}^K {\{ {\mathcal{L}}_{ELBO}^c({\bf x}_{j} ;\theta ,\omega )\} } \Big\} } \,,
\label{mixtureCriterion}
\end{aligned}
\end{equation}
where ${\bf x}_j$ is the $j$-th sample from the joint memory $\mathcal{M}^e_i \cup \mathcal{M}^l_i$. $N'= n_i^e + n_i^l$ and ${\rm R}_i$ is the loss evaluated on all memorized samples using the mixture model at the training step $t_i$. ${\rm R}_{last}$ is the most recent loss value. The pseudocode of the algorithm is provided in Appendix-H from SM.

\section{Experiments}
\label{sec:experiment}
\subsection{Experiment setting and datasets} 

\noindent \textBF{Datasets.}
For the Log-likelihood evaluation, we have the following settings: 
\begin{inparaenum}[1)]
\item \textBF{Split MNIST/Fashion.} Split MNIST \cite{MNIST} into ten parts according to the category information and create a data stream by collecting these parts in a class-incremental way. This is also done for Fashion database;
\item \textBF{Split MNIST-Fashion.} Combine Split MNIST and Split Fashion into a data stream; \item \textBF{Cross-Domain.} Combine Split MNIST-Fashion and unsorted samples from OMNIGLOT \cite{Omniglot}. 
\end{inparaenum} We adapt CIFAR10 \cite{CIFAR10} and Tiny-ImageNet \cite{TinyImageNet} for the generative modelling task. Similar to Split MNIST, we divide CIFAR10 and Tiny-ImageNet into ten parts, namely Split CIFAR10 and Split Tiny-ImageNet, respectively. The details of dataset, hyperparameter and network architecture are provided in Appendix-H.1 of SM.

\noindent \textBF{Evaluation Criteria.}
We use the Inception Score (IS) \cite{InceptScore} and Fréchet Inception Distance (FID) \cite{FID}  for the evaluation of reconstruction quality. For the density estimation task, we estimate the real sample log-likelihood by using IWVAE bound \cite{IWVAE}, as in Eq.~\eqref{IWVAE_loss}, considering 5000 importance samples.

\noindent \textBF{Baseline.} We introduce several baselines used in experiments:
\begin{inparaenum}[1)]
\item VAE-ELBO-OCM~: We train a single VAE model with ELBO using the proposed OCM.
\item VAE-IWVAE50-OCM~: We train a single VAE model with IWVAE using the proposed OCM where the number of importance samples is 50.
\item VAE-ELBO-Random~: We train a single VAE model with a memory that randomly removes samples when it reaches the maximum memory size.
\item Dynamic-ELBO-OCM~: We train a mixture model with ELBO using the proposed OCM.
\item CNDPM \cite{NeuralDirichelt}~: CNDPM uses Dirichlet process for the expansion of the mixture system; \item LIMix \cite{LifelongInfinite}~: We assign an episodic memory with a fixed buffer size for the LIMix model used for TFCL. The maximum number of components for various models is set to 30 to avoid memory overload.
\end{inparaenum}

\makeatletter\def\@captype{table}\makeatother
{
\centering
\begin{minipage}{0.63\textwidth}
\centering
	\caption{The classification accuracy of five indepdnent runs for various models on three datasets.}
	    \scriptsize
\setlength{\tabcolsep}{0.6mm}{\begin{tabular}{@{}l cc cc c@{} } 
\toprule  
\textBF{Methods}   &\textBF{Split MNIST}&\textBF{Split CIFAR10} &\textBF{Split CIFAR100} \\
\midrule 
	  finetune*&19.75 $\pm$ 0.05&18.55 $\pm$ 0.34&3.53 $\pm$ 0.04 \\
	  GEM* \cite{GradientEpisodic} &93.25 $\pm$ 0.36&24.13 $\pm$ 2.46&11.12 $\pm$ 2.48 \\
	  iCARL* \cite{icarl} &83.95 $\pm$ 0.21&37.32 $\pm$ 2.66&10.80 $\pm$ 0.37 \\
	  reservoir* \cite{reservoir} &92.16 $\pm$ 0.75&42.48 $\pm$ 3.04&19.57 $\pm$ 1.79 \\
	 MIR* \cite{OnlineContinualLearning} &93.20 $\pm$ 0.36&42.80 $\pm$ 2.22&20.00 $\pm$ 0.57 \\
	 GSS* \cite{GradientLifelong} &92.47 $\pm$ 0.92&38.45 $\pm$ 1.41&13.10 $\pm$ 0.94 \\
	CoPE-CE* \cite{ContinualPrototype}&91.77 $\pm$ 0.87&39.73 $\pm$ 2.26&18.33 $\pm$ 1.52 \\
	CoPE* \cite{ContinualPrototype} &93.94 $\pm$ 0.20&48.92 $\pm$ 1.32&21.62 $\pm$ 0.69 \\
	CURL* \cite{LifelongUnsupervisedVAE} &92.59 $\pm$ 0.66&-&- \\
	  CNDPM* \cite{NeuralDirichelt} &93.23 $\pm$ 0.09&45.21 $\pm$ 0.18&20.10 $\pm$ 0.12 \\
	 \hline
	 \hline
	   Dynamic-OCM& \textBF{94.02} $\pm$ \textBF{0.23} & \textBF{49.16}  $\pm$ \textBF{1.52} & \textBF{21.79} $\pm$ \textBF{0.68}\\
\bottomrule 
\end{tabular}
\label{classification}
}
\;\;\;\;\;\;\;\;\;\;
\end{minipage}
\makeatletter\def\@captype{table}\makeatother
\begin{minipage}{.35\textwidth}
\centering
	    \scriptsize
	   \renewcommand\arraystretch{1.2}
		\caption{IS and FID on ImageNet database.}
\setlength{\tabcolsep}{0.3mm}{\begin{tabular}{@{}l cc c c@{} } 
\toprule 
\textBF{Model}  &  {\em IS} &{\em FID}  \\
			 		\midrule
		MVAE-Gau \cite{DeepMixtureVAE}  & {\bf 6.84} &/ \\
		MVAE-Gau fixed \cite{DeepMixtureVAE}  &6.30&/ \\
		MVAE-GS \cite{DeepMixtureVAE}  &6.52&/ \\
		MSVI \cite{MSVI}  &6.12&/ \\
	    InfoVAE \cite{InfoVAE}  &6.14 &/ \\
	    $\beta$-VAE \cite{baeVAE}  &5.05 &/ \\
	     VAE \cite{VAE}    &5.46&/ \\
 	     MAE \cite{MutualVAE}   &5.87 &/ \\
 	   	 VAE-ELBO-Random&3.15&145.36\\
 	   \hline
	 \hline
	 VAE-ELBO-OCM & 3.36&133.23 \\
 	  	\bottomrule
	\end{tabular}
	\label{ImageNet}
	}
\end{minipage}
}

\subsection{Log-likelihood evaluation}
\label{subsec:logLikelihood}
In this section, we implement each VAE model or component by using the Bernoulli decoder. All datasets are binarized according to the setting from \cite{IWVAE}. The results for Split MNIST, Split Fashion, Split MNIST-Fashion and Cross-domain are provided in Tables~\ref{tab:logLikelihood} and \ref{tab:logLikelihood2}, where ``Memory'' represents the number of samples $N^l$ in LTM. The proposed OCM can improve the performance on the density estimation tasks even when using a small memory size compared to the random selection approach. Additionally, the expansion mechanism combined with the proposed OCM can further improve the performance with a reasonable memory use, especially when learning multiple datasets (Split MNIST-Fashion and Cross-Domain). We also find that the use of IWVAE bound (Eq.~\eqref{IWVAE_loss}) into the proposed OCM can also improve the performance on a single dataset. To compare with the expansion models, such as LIMix and CNDPM, a single model with OCM outperforms these models by using a few more stored samples such as 2.0K for LTM and 0.5K for STM vs 2.0K for LIMix and CNDPM, in Cross-Domain experiments. However, OCM with the expansion mechanism outperforms LIMix and CNDPM by using fewer mixture components.

\subsection{Evaluation of the reconstruction quality}

To evaluate the reconstruction quality, we use $\beta$-VAE loss \cite{baeVAE} where $\beta = 0.01$ for all models in order to avoid the over-regularization issue \cite{VAE_Prior}. We report the IS and FID scores for the reconstruction quality in Table~\ref{SplitCIFAR10_TinyImageNet}. We can observe that the proposed OCM with the expansion mechanism outperforms other baselines. The IS and FID for Tiny-ImageNet are reported in Appendix-H.3 from SM.

We also explore training a single VAE with OCM for learning ImageNet \cite{ImageNet} under TFCL where the batch size is 64. The maximum size for STM and LTM is set to 512 and 2048, respectively, to avoid increasing the computational cost. We follow the settings from \cite{DeepMixtureVAE}, as described in Appendix-H.3 from SM, after resizing all images to $64 \times 64$ pixels. The FID and IS results are provided in Table~\ref{ImageNet} and the results of all baselines (training on a single dataset) are cited from \cite{DeepMixtureVAE}. The visual results are shown in Fig.~\ref{Reco_ImageNet} where we can observe that the reconstruction of VAE-ELBO-Random is blurred when compared with VAE-ELBO-OCM. These results show that the proposed OCM outperforms the random selection approach in the large-scale dataset under TFCL.

\subsection{Classification task}

The proposed approach is mainly used in unsupervised learning. We also show that OCM can be used in classification tasks when we train a classifier with OCM on the labelled dataset. We adapt the setting and network architecture from \cite{ContinualPrototype} with a batch size of 10 and the memory size for Split MNIST, Split CIFAR10 and Split CIFAR100 is limited to 2K, 1K and 5K, respectively. We report the results in Table~\ref{classification} where `*' means that the result is cited from \cite{ContinualPrototype}. The additional information about baselines and the proposed Dynamic-OCM is provided in Appendix-H.2 of SM. The number of required parameters is provided in Appendix-H.6 of SM. These results show that the proposed OCM outperforms the state-of-the-art methods in the classification task using fewer parameters.

\vspace{-10pt}
\subsection{Ablation study and theoretical results}
\label{sec:ablation}

A full ablation study is performed including testing the configuration for the threshold $\lambda$ from Eq~\eqref{criterion}, STM memory size, batch size and $\lambda_2$ from Eq.~\eqref{mixtureCriterion}. We also provide the empirical results for the theoretical analysis. These ablation results and their analysis are provided in Appendix-H.4 from SM.

\vspace{-10pt}
\section{Conclusion}

We introduce a new theoretical framework for providing insights into the forgetting behaviour of deep models based on VAEs under TFCL. The theoretical analysis demonstrates that ensuring a diversity of data in the pre-training memory is crucial for relieving forgetting in continuous learning systems. Inspired by this result, we propose the Online Cooperative Memorization (OCM) that does not require any supervised signals and therefore can be used in an unsupervised fashion. The empirical results demonstrate the effectiveness of the proposed OCM method.

\clearpage
%
%
\bibliographystyle{splncs04}
\bibliography{ECCV_Version.bbl}

\begin{thebibliography}{10}
\providecommand{\url}[1]{\texttt{#1}}
\providecommand{\urlprefix}{URL }
\providecommand{\doi}[1]{https://doi.org/#1}

\bibitem{InfVAE}
Abbasnejad, E., Dick, M., van~der Hengel, A.: Infinite variational autoencoder
  for semi-supervised learning. In: Proc. of IEEE Conf. on Computer Vision and
  Pattern Recognition (CVPR). pp. 5888--5897 (2017)

\bibitem{Lifelong_VAE}
Achille, A., Eccles, T., Matthey, L., Burgess, C., Watters, N., Lerchner, A.,
  Higgins, I.: Life-long disentangled representation learning with cross-domain
  latent homologies. In: Proc. Advances in Neural Inf. Proc. Systems (NeurIPS).
  pp. 9873--9883 (2018)

\bibitem{GradientLifelong}
Aljundi, R., Lin, M., Goujaud, B., Bengio, Y.: Gradient based sample selection
  for online continual learning. In: Advances Neural Information Processing
  Systems (NeurIPS). vol.~33, pp. 11817--11826 (2019)

\bibitem{OnlineContinualLearning}
Aljundi, R., Belilovsky, E., Tuytelaars, T., Charlin, L., Caccia, M., Lin, M.,
  Page-Caccia, L.: Online continual learning with maximal interfered retrieval.
  In: Advances in Neural Information Processing Systems (NeurIPS). vol.~33, pp.
  11872--11883 (2019)

\bibitem{taskFree_CL}
Aljundi, R., Kelchtermans, K., Tuytelaars, T.: Task-free continual learning.
  In: Proc. of IEEE/CVF Conf. on Computer Vision and Pattern Recognition. pp.
  11254--11263 (2019)

\bibitem{RainbowMemory}
Bang, J., Kim, H., Yoo, Y., Ha, J.W., Choi, J.: Rainbow memory: Continual
  learning with a memory of diverse samples. In: Proceedings of the IEEE/CVF
  Conference on Computer Vision and Pattern Recognition. pp. 8218--8227 (2021)

\bibitem{MutualInformation}
Belghazi, M.I., Baratin, A., Rajeshwar, S., Ozair, S., Bengio, Y., Courville,
  A., Hjelm, D.: Mutual information neural estimation. In: Proc. Inter.
  Conference on Machine Learning (ICML), vol. PMLR 80. pp. 531--540 (2018)

\bibitem{WVAE2}
Bousquet, O., Gelly, S., Tolstikhin, I., Simon-Gabriel, C.J., Schoelkopf, B.:
  From optimal transport to generative modeling: the {VEGAN} cookbook. arXiv
  preprint arXiv:1705.07642  (2017)

\bibitem{IWVAE}
Burda, Y., Grosse, R., Salakhutdinov, R.: Importance weighted autoencoders.
  arXiv preprint arXiv:1509.00519  (2015)

\bibitem{TinyLifelong}
Chaudhry, A., Rohrbach, M., Elhoseiny, M., Ajanthan, T., Dokania, P., Torr,
  P.H.S., Ranzato, M.: On tiny episodic memories in continual learning. arXiv
  preprint arXiv:1902.10486  (2019)

\bibitem{VAE_symmetric}
Chen, L., Dai, S., Pu, Y., Li, C., Su, Q., Carin, L.: Symmetric variational
  autoencoder and connections to adversarial learning. In: Proc. Int. Conf. on
  Artificial Intel. and Statistics (AISTATS) 2018, vol. PMLR 84. pp. 661--669
  (2018)

\bibitem{OptimalTransport}
Courty, N., Flamary, R., Tuia, D., Rakotomamonjy, A.: Optimal transport for
  domain adaptation. IEEE Trans. on Pattern Analysis and Machine Intelligence
  \textbf{39}(9),  1853--1865 (2016)

\bibitem{ContinualPrototype}
De~Lange, M., Tuytelaars, T.: Continual prototype evolution: Learning online
  from non-stationary data streams. In: Proc. of the IEEE/CVF Int. Conference
  on Computer Vision (ICCV). pp. 8250--8259 (2021)

\bibitem{BooVAE}
Egorov, E., Kuzina, A., Burnaev, E.: {BooVAE}: Boosting approach for continual
  learning of {VAE}. Advances in Neural Information Processing Systems
  (NeurIPS)  \textbf{35},  17889--17901 (2021)

\bibitem{KernelMethods}
Fang, P., Harandi, M., Petersson, L.: Kernel methods in hyperbolic spaces. In:
  Proc. of the IEEE/CVF Int. Conference on Computer Vision (ICCV). pp.
  10665--10674 (2021)

\bibitem{UnbalancedOT}
Fatras, K., S{\'e}journ{\'e}, T., Flamary, R., Courty, N.: Unbalanced minibatch
  optimal transport; applications to domain adaptation. In: Int. Conf. on
  Machine Learning (ICML), vol. PMLR 139. pp. 3186--3197 (2021)

\bibitem{KLForMixture}
Goldberger, J., Gordon, S., Greenspan, H., et~al.: An efficient image
  similarity measure based on approximations of kl-divergence between two
  gaussian mixtures. In: Proc. IEEE Int. Conf. on Computer Vision (ICCV).
  vol.~3, pp. 487--493 (2003)

\bibitem{GAN}
Goodfellow, I., Pouget-Abadie, J., Mirza, M., Xu, B., Warde-Farley, D., Ozair,
  S., Courville, A., Bengio, Y.: Generative adversarial nets. In: Proc.
  Advances in Neural Inf. Proc. Systems (NIPS). pp. 2672--2680 (2014)

\bibitem{FID}
Heusel, M., Ramsauer, H., Unterthiner, T., Nessler, B., Hochreiter, S.: Gans
  trained by a two time-scale update rule converge to a local {N}ash
  equilibrium. In: Proc. Advances in Neural Information Processing Systems
  (NIPS). pp. 6626--6637 (2017)

\bibitem{baeVAE}
Higgins, I., Matthey, L., Pal, A., Burgess, C., Glorot, X., Botvinick, M.,
  Mohamed, S., Lerchner, A.: $\beta$-{VAE}: Learning basic visual concepts with
  a constrained variational framework. In: Proc. Int. Conf. on Learning
  Representations (ICLR) (2017)

\bibitem{Distilling_nets}
Hinton, G., Vinyals, O., Dean, J.: Distilling the knowledge in a neural
  network. In: Proc. NIPS Deep Learning Workshop, arXiv preprint
  arXiv:1503.02531 (2014)

\bibitem{LSTM_Time}
Hua, Y., Zhao, Z., Li, R., Chen, X., Liu, Z., Zhang, H.: Deep learning with
  long short-term memory for time series prediction. IEEE Communications
  Magazine  \textbf{57}(6),  114--119 (2019)

\bibitem{LessForgetting}
Jung, H., Ju, J., Jung, M., Kim, J.: Less-forgetting learning in deep neural
  networks. arXiv preprint arXiv:1607.00122  (2016)

\bibitem{OptimalTransportBook}
Kantorovitch, L.: On the translocation of masses. Management science
  \textbf{5}(1), ~1--4 (1958)

\bibitem{VAE}
Kingma, D.P., Welling, M.: Auto-encoding variational {B}ayes. arXiv preprint
  arXiv:1312.6114  (2013)

\bibitem{EWC}
Kirkpatrick, J., Pascanu, R., Rabinowitz, N., Veness, J., Desjardins, G., Rusu,
  A.A., Milan, K., Quan, J., Ramalho, T., Grabska-Barwinska, A., Hassabis, D.,
  Clopath, C., Kumaran, D., Hadsell, R.: Overcoming catastrophic forgetting in
  neural networks. Proc. of the National Academy of Sciences (PNAS)
  \textbf{114}(13),  3521--3526 (2017)

\bibitem{OptimalCL}
Knoblauch, J., Husain, H., Diethe, T.: Optimal continual learning has perfect
  memory and is {NP}-hard. In: Proc. International Conference on Machine
  Learning (ICML), vol PMLR 119. pp. 5327--5337 (2020)

\bibitem{CIFAR10}
Krizhevsky, A., Hinton, G.: Learning multiple layers of features from tiny
  images. Tech. rep. (2009)

\bibitem{ImageNet}
Krizhevsky, A., Sutskever, I., Hinton, G.E.: Imagenet classification with deep
  convolutional neural networks. In: Advances in Neural Inf. Proc. Systems
  (NIPS). pp. 1097--1105 (2012)

\bibitem{MSVI}
Kurle, R., G{\"u}nnemann, S., van~der Smagt, P.: Multi-source neural
  variational inference. In: Proc. of AAAI Conf. on Artificial Intelligence.
  vol.~33, pp. 4114--4121 (2019)

\bibitem{Omniglot}
Lake, B.M., Salakhutdinov, R., Tenenbaum, J.B.: Human-level concept learning
  through probabilistic program induction. Science  \textbf{350}(6266),
  1332--1338 (2015)

\bibitem{TinyImageNet}
Le, Y., Yang, X.: Tiny imagenet visual recognition challenge. CS 231N
  \textbf{7}(7), ~3 (2015)

\bibitem{MNIST}
LeCun, Y., Bottou, L., Bengio, Y., Haffner, P.: Gradient-based learning applied
  to document recognition. Proc. of the IEEE  \textbf{86}(11),  2278--2324
  (1998)

\bibitem{CLTeacherStudent}
Lee, S., Goldt, S., Saxe, A.: Continual learning in the teacher-student setup:
  Impact of task similarity. In: International Conference on Machine Learning
  (ICML), vol. PMLR 139. pp. 6109--6119 (2021)

\bibitem{NeuralDirichelt}
Lee, S., Ha, J., Zhang, D., Kim, G.: A neural {D}irichlet process mixture model
  for task-free continual learning. In: Proc. Int. Conf. on Learning
  Representations (ICLR), arXiv preprint arXiv:2001.00689 (2020)

\bibitem{Lwf}
Li, Z., Hoiem, D.: Learning without forgetting. IEEE Trans. on Pattern Analysis
  and Machine Intelligence  \textbf{40}(12),  2935--2947 (2017)

\bibitem{WGAN_Transport}
Liu, H., Gu, X., Samaras, D.: Wasserstein {GAN} with quadratic transport cost.
  In: Proc. of the IEEE/CVF Int. Conf. on Computer Vision (ICCV). pp.
  4832--4841 (2019)

\bibitem{ItoI_network}
Liu, M.Y., Breuel, T., Kautz, J.: Unsupervised image-to-image translation
  networks. In: Advances in Neural Information Processing Systems. pp. 700--708
  (2017)

\bibitem{GradientEpisodic}
Lopez-Paz, D., Ranzato, M.: Gradient episodic memory for continual learning.
  In: Advances in Neural Information Processing Systems. pp. 6467--6476 (2017)

\bibitem{MutualVAE}
Ma, X., Zhou, C., Hovy, E.: {MAE}: Mutual posterior-divergence regularization
  for variational autoencoders. In: Proc. Int. Conf. on Learning
  Representations (ICLR), arXiv preprint arXiv:1901.01498 (2019)

\bibitem{VCL}
Nguyen, C.V., Li, Y., Bui, T.D., Turner, R.E.: Variational continual learning.
  In: Proc. of Int. Conf. on Learning Representations (ICLR), arXiv preprint
  arXiv:1710.10628 (2018)

\bibitem{LifeLong_review}
Parisi, G.I., Kemker, R., Part, J.L., Kanan, C., Wermter, S.: Continual
  lifelong learning with neural networks: A review. Neural Networks
  \textbf{113},  54--71 (2019)

\bibitem{CL_TradeOff}
Raghavan, K., Balaprakash, P.: Formalizing the generalization-forgetting
  trade-off in continual learning. Advances in Neural Information Processing
  Systems  \textbf{34} (2021)

\bibitem{GenerativeLifelong}
Ramapuram, J., Gregorova, M., Kalousis, A.: Lifelong generative modeling. In:
  Proc. Int. Conf. on Learning Representations (ICLR), arXiv preprint
  arXiv:1705.09847 (2017)

\bibitem{LifelongUnsupervisedVAE}
Rao, D., Visin, F., Rusu, A.A., Teh, Y.W., Pascanu, R., Hadsell, R.: Continual
  unsupervised representation learning. In: Advances Neural Inf. Processing
  Systems (NeurIPS). pp. 7645--7655 (2019)

\bibitem{icarl}
Rebuffi, S.A., Kolesnikov, A., Sperl, G., Lampert, C.H.: {iCaRL}: Incremental
  classifier and representation learning. In: Proc. of the IEEE Conf. on
  Computer Vision and Pattern Recognition (CVPR). pp. 2001--2010 (2017)

\bibitem{LifeLong_combination}
Ren, B., Wang, H., Li, J., Gao, H.: Life-long learning based on dynamic
  combination model. Applied Soft Computing  \textbf{56},  398--404 (2017)

\bibitem{InceptScore}
Salimans, T., Goodfellow, I., Zaremba, W., Cheung, V., Radford, A., Chen, X.:
  Improved techniques for training {GANs}. In: Proc. Advances in Neural Inf.
  Proc. Systems (NIPS). pp. 2234--2242 (2016)

\bibitem{Generative_replay}
Shin, H., Lee, J.K., Kim, J., Kim, J.: Continual learning with deep generative
  replay. In: Advances in Neural Inf. Proc. Systems (NIPS). pp. 2990--2999
  (2017)

\bibitem{ImportanceWeightedHierarchical}
Sobolev, A., Vetrov, D.: Importance weighted hierarchical variational
  inference. In: Advances in Neural Information Processing Systems (NeurIPS).
  vol.~33 (2019)

\bibitem{VAE_Prior}
Takahashi, H., Iwata, T., Yamanaka, Y., Yamada, M., Yagi, S.: Variational
  autoencoder with implicit optimal priors. In: Proc. of the AAAI Conference on
  Artificial Intelligence. vol.~33, pp. 5066--5073 (2019)

\bibitem{LayerwiseCL}
Tang, S., Chen, D., Zhu, J., Yu, S., Ouyang, W.: Layerwise optimization by
  gradient decomposition for continual learning. In: Proc. of the IEEE/CVF
  Conference on Computer Vision and Pattern Recognition (CVPR). pp. 9634--9643
  (2021)

\bibitem{FunctionalRegularisation}
Titsias, M.K., Schwarz, J., Matthews, A.G.d.G., Pascanu, R., Teh, Y.W.:
  Functional regularisation for continual learning with {G}aussian processes.
  In: Proc. Int. Conf. on Learning Represenations (ICLR), arXiv preprint
  arXiv:1901.11356 (2019)

\bibitem{WVAE}
Tolstikhin, I., Bousquet, O., Gelly, S., Schoelkopf, B.: Wasserstein
  auto-encoders. In: Int. Conf. on Learning Representations (ICLR), arXiv
  preprint arXiv:1711.01558 (2018)

\bibitem{reservoir}
Vitter, J.S.: Random sampling with a reservoir. ACM Transactions on
  Mathematical Software (TOMS)  \textbf{11}(1),  37--57 (1985)

\bibitem{NullSpaceCL}
Wang, S., Li, X., Sun, J., Xu, Z.: Training networks in null space of feature
  covariance for continual learning. In: Proc. of the IEEE/CVF Conf. on
  Computer Vision and Pattern Recognition (CVPR). pp. 184--193 (2021)

\bibitem{LifelongTeacherStudent}
Ye, F., Bors, A.: Lifelong teacher-student network learning. IEEE Trans. on
  Pattern Analysis and Machine Intelligence  (2021).
  \doi{10.1109/TPAMI.2021.3092677}

\bibitem{LifelongVAEGAN}
Ye, F., Bors, A.G.: Learning latent representations across multiple data
  domains using lifelong {VAEGAN}. In: Proc. European Conf. on Computer Vision
  (ECCV), vol. LNCS 12365. pp. 777--795 (2020)

\bibitem{Lifelonginterpretable}
Ye, F., Bors, A.G.: Lifelong learning of interpretable image representations.
  In: Proc. Int. Conf. on Image Processing Theory, Tools and Applications
  (IPTA). pp.~1--6 (2020)

\bibitem{MixtureOfVAEs}
Ye, F., Bors, A.G.: Mixtures of variational autoencoders. In: 2020 Tenth
  International Conference on Image Processing Theory, Tools and Applications
  (IPTA). pp.~1--6 (2020)

\bibitem{DeepMixtureVAE}
Ye, F., Bors, A.G.: Deep mixture generative autoencoders. IEEE Transactions on
  Neural Networks and Learning Systems pp. 1--15 (2021).
  \doi{10.1109/TNNLS.2021.3071401}

\bibitem{InfoVAEGAN_conference}
Ye, F., Bors, A.G.: Infovaegan: Learning joint interpretable representations by
  information maximization and maximum likelihood. In: Proc. IEEE Int. Conf. on
  Image Processing (ICIP). pp. 749--753 (2021).
  \doi{10.1109/ICIP42928.2021.9506169}

\bibitem{JontLatentVAEs}
Ye, F., Bors, A.G.: Learning joint latent representations based on information
  maximization. Information Sciences  \textbf{567},  216--236 (2021)

\bibitem{LifelongInfinite}
Ye, F., Bors, A.G.: Lifelong infinite mixture model based on knowledge-driven
  {D}irichlet process. In: Proc. of the IEEE Int. Conf. on Computer Vision
  (ICCV) (2021)

\bibitem{LifelongMixuteOfVAEs}
Ye, F., Bors, A.G.: Lifelong mixture of variational autoencoders. IEEE
  Transactions on Neural Networks and Learning Systems pp. 1--14 (2021).
  \doi{10.1109/TNNLS.2021.3096457}

\bibitem{LifelongTwin}
Ye, F., Bors, A.G.: Lifelong twin generative adversarial networks. In: Proc.
  IEEE Int. Conf. on Image Processing (ICIP). pp. 1289--1293 (2021)

\bibitem{Learning_Evolved}
Ye, F., Bors, A.G.: Learning an evolved mixture model for task-free continual
  learning (2022)

\bibitem{LifelongGraph}
Ye, F., Bors, A.G.: Lifelong generative modelling using dynamic expansion graph
  model. In: AAAI on Artificial Intelligence. AAAI Press (2022)

\bibitem{LifelongGAN}
Zhai, M., Chen, L., Tung, F., He, J., Nawhal, M., Mori, G.: Lifelong {GAN}:
  Continual learning for conditional image generation. In: Proc. of the
  IEEE/CVF Int. Conf. on Computer Vision (ICCV). pp. 2759--2768 (2019)

\bibitem{InfoVAE}
Zhao, S., Song, J., Ermon, S.: Info{VAE}: Balancing learning and inference in
  variational autoencoders. In: Proc. AAAI Conf. on Artif. Intel. vol.~33, pp.
  5885--5892 (2019)

\end{thebibliography}
\end{document}